\documentclass{article}

\PassOptionsToPackage{numbers, compress}{natbib}

\usepackage[final]{neurips_2018}

\usepackage[utf8]{inputenc} 
\usepackage[T1]{fontenc}    
\usepackage{hyperref}       
\usepackage{url}            
\usepackage{booktabs}       
\usepackage{amsfonts}       
\usepackage{nicefrac}       
\usepackage{microtype}      

\usepackage{graphicx}
\usepackage{algorithm}
\usepackage{algorithmic}
\usepackage{amsmath}
\DeclareMathOperator*{\argmaxA}{argmax}
\usepackage{caption}
\usepackage{kotex}

\usepackage{color}
\newcommand{\rt}{\textcolor[rgb]{0,0,0}}
\newcommand{\ot}{\textcolor[rgb]{0,0,0}}
\newcommand{\bt}{\textcolor[rgb]{0,0,0}}
\newcommand{\gt}{\textcolor[rgb]{0,0,0}}

\title{Answerer in Questioner's Mind:\\ Information Theoretic Approach to \\ Goal-Oriented Visual Dialog}

\author{
  Sang-Woo Lee$^1$\thanks{Work carried out at Seoul National University} \ , Yu-Jung Heo$^2$, and Byoung-Tak Zhang$^{2,3}$\\
  \\
  Clova AI Research, Naver Corp$^1$\\
  Seoul National University$^2$\\
  \rt{Surromind Robotics}$^3$
}

\begin{document}

\maketitle

\begin{abstract}
Goal-oriented dialog has been given attention due to its numerous applications in artificial intelligence.
Goal-oriented dialogue tasks occur when a questioner asks an action-oriented question and an answerer responds with the intent of letting the questioner know a correct action to take. 
To ask the adequate question, deep learning and reinforcement learning have been recently applied. 
However, these approaches struggle to find a competent recurrent neural questioner, owing to the complexity of learning a series of sentences.
Motivated by theory of mind, we propose ``Answerer in Questioner's Mind'' (AQM), a novel information theoretic algorithm for goal-oriented dialog. 
With AQM, a questioner asks and infers based on an approximated probabilistic model of the answerer.
The questioner figures out the answerer’s intention via selecting a plausible question by explicitly calculating the information gain of the candidate intentions and possible answers to each question.
We test our framework on two goal-oriented visual dialog tasks: ``MNIST Counting Dialog'' and ``GuessWhat?!''.
In our experiments, AQM outperforms comparative algorithms by a large margin.
\end{abstract}

\section{Introduction}
Goal-oriented dialog is a classical artificial intelligence problem \rt{that needs to be addressed for} digital personal assistants, order-by-phone tools, and online customer service centers.
Goal-oriented dialog occurs when a questioner asks an action-oriented question and an answerer responds with the intent of letting the questioner know a correct action to take.
Significant research on goal-oriented dialog has tackled this problem \rt{using from the rule-based approach to the} \ot{end-to-end neural} \rt{approach \cite{lemon2006,williams2007,bordes2017}.}

Motivated by the achievement of neural chit-chat dialog research \cite{vinyals2015a}, recent studies on goal-oriented dialogs have utilized deep learning, using massive data to train their neural networks in self-play environments. 
In this setting, two machine agents are trained to make a dialog to achieve the goal of the task in a cooperative way \cite{lazaridou2018,de2017,das2017a}.
Many researchers attempted to solve goal-oriented dialog tasks by using the deep supervised learning (deep SL) approach \cite{wen2016} based on seq2seq models \cite{cho2014} or the deep reinforcement learning (deep RL) approach utilizing rewards obtained from the result of the dialog \cite{zhao2016,li2017}.
However, these methods struggle to find a competent RNN model that uses back-propagation, owing to the complexity of learning a series of sentences. These algorithms tend to generate redundant sentences, making generated dialogs inefficient \cite{kim2017,das2017b}.

Our idea to deal with goal-oriented dialog is motivated by theory of mind \cite{premack1978}, the ability to attribute mental states to others and to understand how our mental states are different. 
In this approach, an agent considers what the collaborator, the opposite agent cooperating in dialog, will respond by using an explicit approximated model of the collaborator.
If one wishes to efficiently convey information to the other, it is best to converse in a way that maximizes the other's understanding \cite{bruner1981}. 
For our method, we consider the mind to be beyond a part of mental states (e.g., belief, intent, knowledge). The mind is the probabilistic distribution of the model of the collaborator itself.

\begin{figure}[t] 
\centering
\includegraphics[width=0.48\textwidth]{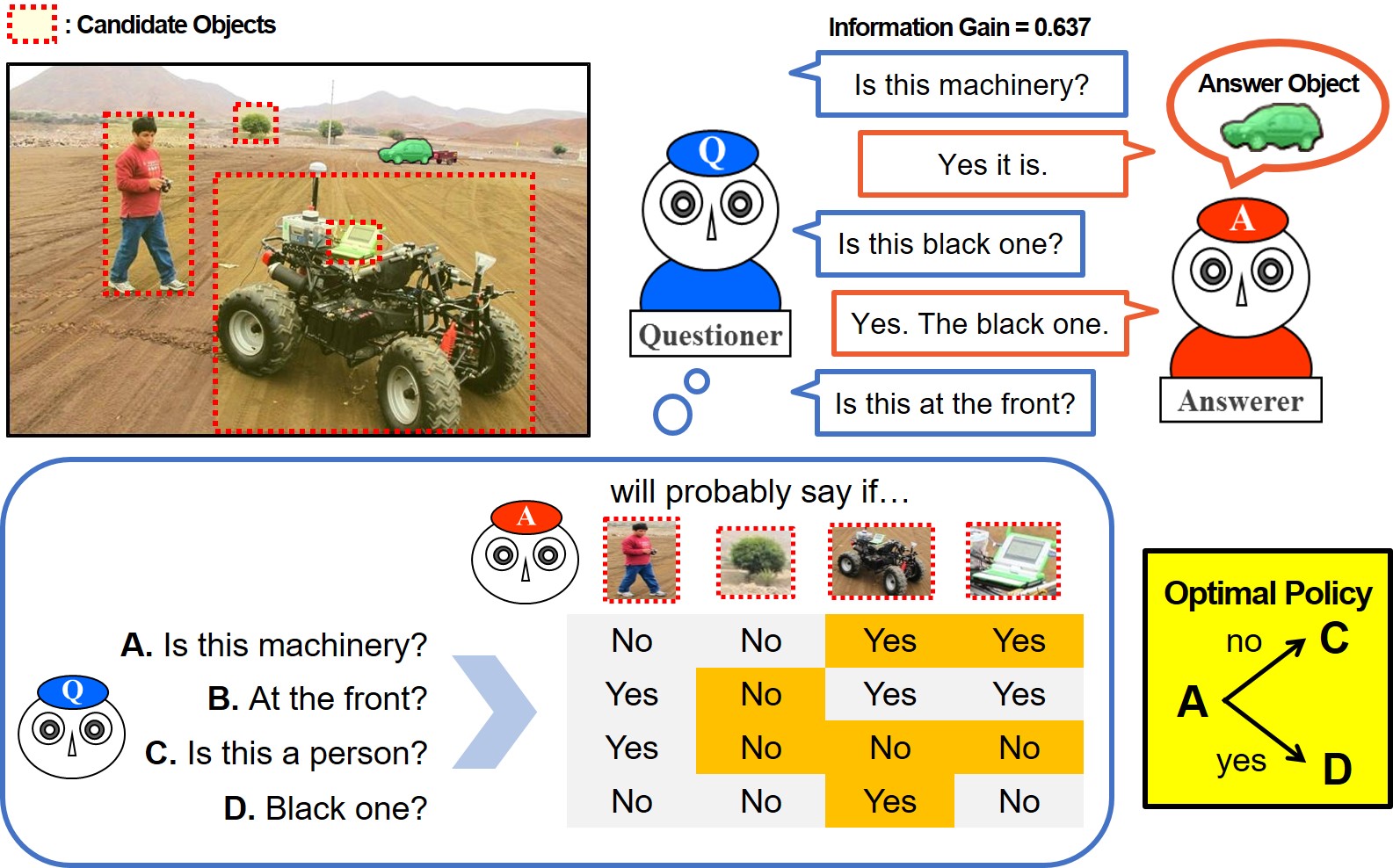}
\hspace{0.4cm}
\includegraphics[width=0.45\textwidth]{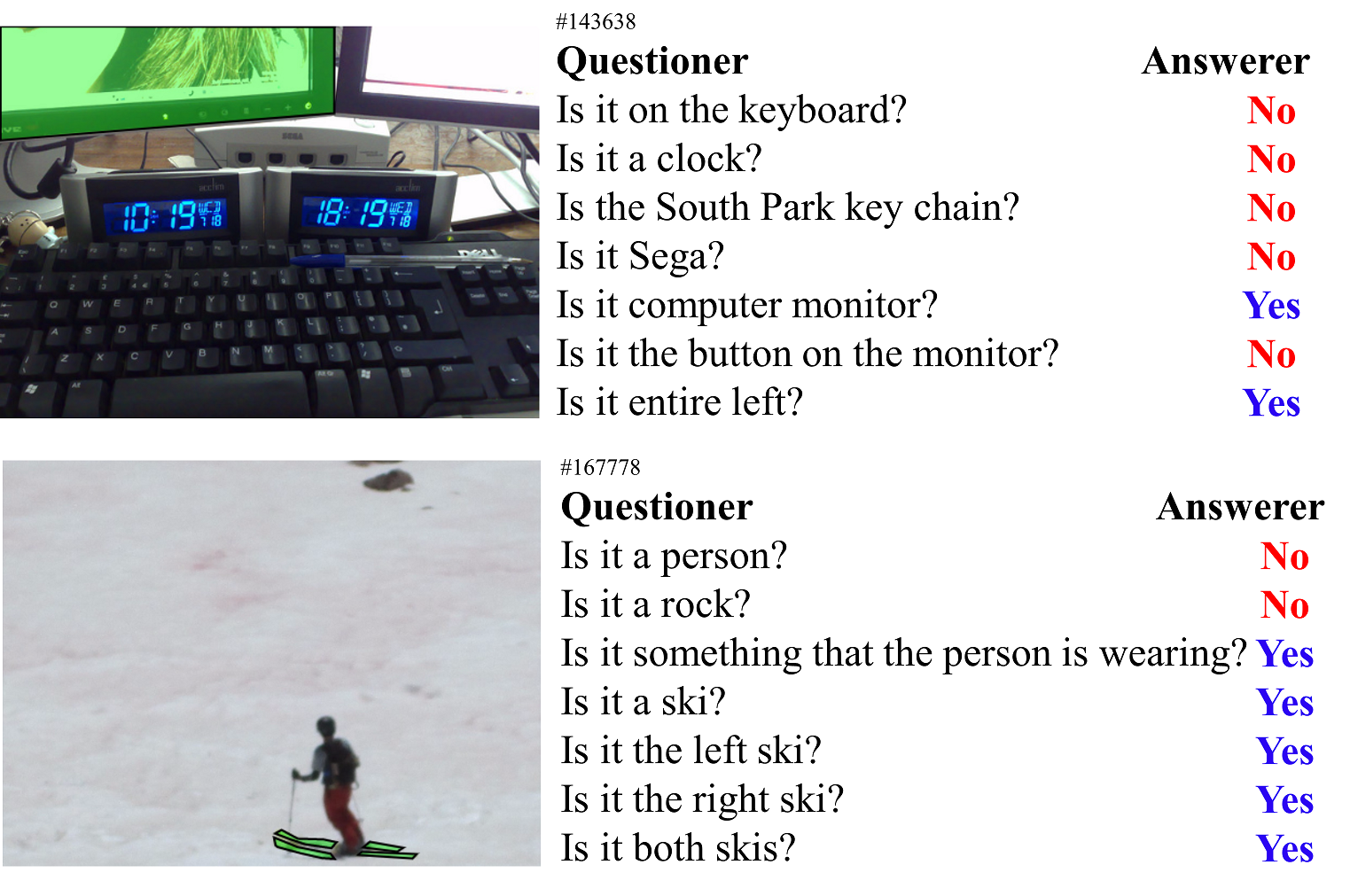}
\caption{(Left) Illustration of an AQM algorithm for goal-oriented visual dialog. AQM makes a decision tree on the image for asking efficient questions. (Right) Examples of the GuessWhat?! game.
The goal of GuessWhat?! is to locate the correct object in the image.
The green mask highlights the correct object. }
\label{fig:icml18_fig12}
\end{figure}

We propose an ``Answerer in Questioner’s Mind'' (AQM) algorithm for goal-oriented dialog (Figure \ref{fig:icml18_fig12} (Left)). AQM allows a questioner to ask appropriate consecutive questions with information theoretic approach \cite{mackay2003,zhang2013}.
AQM's questioner explicitly possesses an approximated model of the answerer. The questioner utilizes the approximated model to calculate the information gain of the candidate answerer's intentions and answers for each question. 
In the example of Figure \ref{fig:icml18_fig12} (Left), the answerer's intention is the correct object highlighted by the green mask.
Using the candidate objects and the approximated answerer model, the questioner makes an efficient question which \rt{splits} out the candidate objects properly.
``Is this machinery?'' is selected in the first turn, because the question
separates the candidate objects evenly and thus has maximum information gain.
In our main experiment, AQM's question generator extracts proper questions from the training data, not generating new questions.
However, in the discussion section, we extend AQM to generate novel questions for the test image.

We test AQM mainly on goal-oriented visual dialog tasks in the self-play environment.
Our main experiment is conducted on ``GuessWhat?!'', a cooperative two-player guessing game on the image (Figure \ref{fig:icml18_fig12} (Right)). AQM achieves an accuracy of 63.63\% in 3 turns and 78.72\% in 10 turns, outperforming deep SL (46.8\% in 5 turns) \cite{de2017} and deep RL (52.3\% in 4.1 turns) \cite{strub2017} algorithms.
Though we demonstrate the performance of our models in visual dialog tasks, our approach can be directly applied to general goal-oriented dialog where there is a non-visual context.

Our main contributions are four folds. 

\begin{itemize}
	\item We propose the AQM, a practical goal-oriented dialog system motivated by theory of mind. The AQM framework is general and not rely on a specific model representation nor a learning method. We compare various types of learning strategy on the model and selecting strategy for the candidate questions.
	\item We test our AQM on two goal-oriented visual dialog tasks, showing that our method outperforms comparative methods.
	\item We use AQM as a tool to understand existing deep learning methods in goal-oriented dialog studies.
	\rt{Section 5.1 and Appendix D include 1) the relationship between the hidden vector in comparative models and the posterior in AQM, 2) the relationship between the objective function of RL and AQM, and 3)} \ot{a point to be considered} \rt{on self-play with RL for making an agent to converse with a human.}
	\item \rt{We} extend AQM to generate questions, in which case AQM can be understood as a way to boost the existing deep learning method \ot{in Section 5.2.} 
\end{itemize}

\section{Previous Works}

\rt{Our study is related to various research fields, including goal-oriented dialog \cite{lemon2006,williams2007,bordes2017,zhao2016,li2017}, language emergence \cite{lazaridou2018,evtimova2018}, the theory of mind \cite{chandrasekaran2017,batali1998,choi2018,hernandez2017}, referring game \cite{choi2018,mao2016}, pragmatics \cite{fried2018,andreas2016,yu2017,monroe2017}, and visual dialog \cite{de2017,das2017a,kim2017,das2017b,strub2017}. In this section, we highlight three topics as below, obverter, opponent modeling, and information gain.}

\textbf{Obverter } 
\rt{Choi et al. recently} applied the obverter technique \cite{batali1998}, motivated by theory of mind, to study language emergence \cite{choi2018}. The task of the study is an image-description-match classification. 
In their experiments, one agent transmitted one sentence for describing an artificial image to the collaborator agent. 
In their study, the obverter technique can be understood as that an agent plays both questioner and answerer, maximizing the consistency between visual and language modules.
Their experimental results showed that their obverter technique generated a word corresponding to a specific object (e.g. `bbbbbbb\{b,d\}' for a blue box). They argued their method could be an alternative to RL-based language emergence systems. Compared to their model, however, AQM uses real images, creates multi-turn dialog, and can be used for general goal-oriented dialog tasks.

\textbf{Opponent Modeling } Studies on opponent modeling have treated simple games with a multi-agent environment where an agent competed with the other \cite{hernandez2017}.
In the study of Foerster et al. \cite{foerster2017}, the agent has the model of the opponent and updates it assuming the opponent will be updated by gradient descent with RL.
They argued modeling opponent could be applied to track the non-stationary behavior of an opponent agent. 
Their model outperformed classical RL methods in simple games, such as tic-tac-toe and rock-paper-scissors. 
On the other hand, AQM applied opponent modeling to a cooperative multi-agent setting. 
We believe that opponent modeling can also be applied to dialog systems in which agents are partially cooperative and partially competitive.

In a broader sense, our study can also be understood as extending these studies to a multi-turn visual dialog, as the referring game is a special case of single-turn visual dialog.

\textbf{Information Gain } AQM's question-generator optimizes information gain using an approximated collaborator model. However, the concept of utilizing information gain in a dialog task is not new for a classical rule-based approach. 
Polifroni and Walker used information gain to build a dialog system for restaurant recommendations \cite{polifroni2006}. 
They made a decision tree using information gain and asked a series of informative questions about restaurant preferences. 
Rothe et al. applied a similar method to generate questions on a simple Battleship game experiment \cite{rothe2017}. 
It is noticeable that they used pre-defined logic to generate questions with information gain criteria to make novel (i.e., not shown in the training dataset) and human-like questions.
Unlike these previous studies, AQM makes a new decision tree for every new context; asking a question in AQM corresponds to constructing a node in decision tree classifier.
In the example of Figure \ref{fig:icml18_fig12} (Left), AQM makes a new decision tree for a test image.
AQM also considers uncertainty by deep learning, and does not require hand-made or domain-specific rules.


\section{Answerer in Questioner's Mind (AQM)}

\textbf{Preliminary}
In our experimental setting, two machine players, a questioner and an answerer, communicate via natural dialog. 
Specifically, there exists a \rt{target} class $c$, which is an answerer’s intention or a goal-action the questioner should perform. 
The answerer knows the class $c$, whereas the questioner does not. 
The goal of the dialog for the questioner is to find the correct class $c$ by asking a series of questions to the answerer. 
The answerer responds the answer to the given question.

We treat $C$, $Q_t$, and $A_t$ as random variables of class, $t$-th question, and $t$-th answer, respectively. $c$, $q_t$, and $a_t$ becomes their single instance. 
In a restaurant scenario example, $q_t$ can be ``Would you like to order?'' or ``What can I do for you?'' $a_t$ can be ``Two coffees, please.'' or ``What's the password for Wi-Fi?'' $c$ can then be ``Receive the order of two hot Americanos.'' or ``Let the customer know the Wi-Fi password.''

\begin{figure}[t] 
\centering
\includegraphics[width=0.90\textwidth]{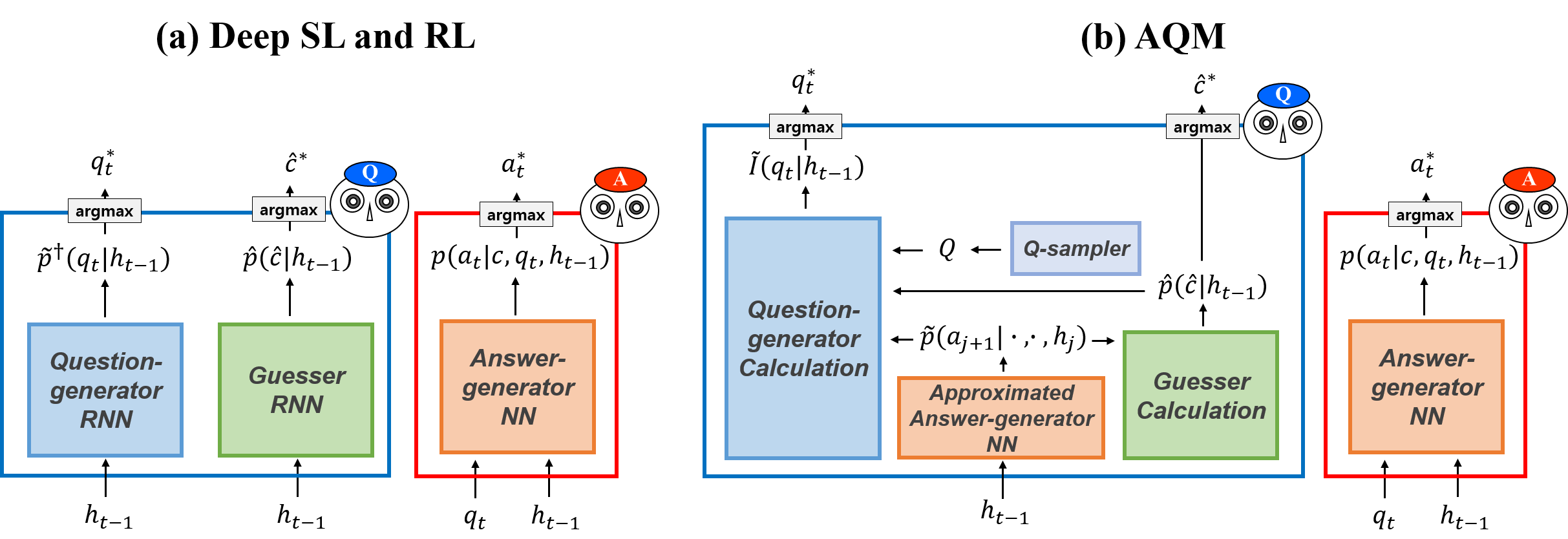}
\caption{Comparative illustration on modules of the existing deep learning framework and an AQM framework. The AQM's guesser computes Equation \ref{eq:eq_posterior} and the AQM's question-generator computes Equation \ref{eq:eq_QG}. $h_{t-1}$ is the previous history $(a_{1:t-1},q_{1:t-1})$.}
\label{fig:nips18_fig8}
\end{figure}

\textbf{Model Structure }
We illustrate the difference between the existing deep learning framework \cite{de2017,strub2017,das2017b} and the proposed AQM framework in Figure \ref{fig:nips18_fig8}.
The answerer systems of two frameworks are the same.
The answerer works as a kind of \ot{simple neural network models for} visual question answering (VQA) and the network is trained on the training data.
On the other hand, the questioner of two frameworks works differently.

In deep SL and RL methods \cite{de2017,strub2017}, the questioner has two RNN-based models, one is the question-generator to generate a question and the other is the guesser to classify a class. 
In AQM, these two RNN-based models are substituted \rt{by} the mathematical calculation of the equation and argmax operation, not the model.
The question-generator selects a question from candidate questions sampled from $Q$-sampler, \rt{which makes a pre-defined question set before the game starts.}
The question-generator calculates information gain $I[C,A_t;q_t,a_{1:t-1},q_{1:t-1}]$ for each question in candidate questions.
$I$ is the information gain or mutual information of the class $C$ and the current answer $A_t$, where the previous history $h_{t-1} = (a_{1:t-1}, q_{1:t-1})$ and the current question $q_t$ are given. 
Note that maximizing information gain is the same \rt{as} minimizing the conditional entropy of class $C$, given a current answer $A_t$.

\begin{equation}
\begin{aligned}
& I[C,A_t;q_t,a_{1:t-1},q_{1:t-1}] \\ 
= & H[C;a_{1:t-1},q_{1:t-1}] - H[C|A_t;q_t,a_{1:t-1},q_{1:t-1}] \\ 
= & \sum_{a_t} \sum_{c} p(c|a_{1:t-1},q_{1:t-1}) p(a_t|c,q_t,a_{1:t-1},q_{1:t-1}) 
\ln \frac{p(a_t|c,q_t,a_{1:t-1},q_{1:t-1})}{p(a_t|q_t,a_{1:t-1},q_{1:t-1})}
\label{eq:eq_infogain}
\end{aligned}
\end{equation}

The guesser calculates the posterior of class $p(c|a_{1:t},q_{1:t}) \propto p(c) \prod_{\rt{j=1}}^{t} p(a_j|c,q_j,a_{1:j-1},q_{1:j-1})$.

\textbf{Calculation }
For the equation of both question-generator and guesser, the answerer's answer distribution $p(a_t|c,q_t,a_{1:t-1},q_{1:t-1})$ is required. 
The questioner has an approximated answer-generator network to make the approximated answer distribution $\tilde{p}(a_t|c,q_t,a_{1:t-1},q_{1:t-1})$, which we refer to as the likelihood $\tilde{p}$.
If $a_t$ is a sentence, the probability of the answer-generator is extracted from the multiplication of the word probability of RNN.

The AQM’s guesser module calculates the posterior of class $c$, $\hat{p}(c|a_{1:t},q_{1:t})$, based on the history $h_t = (a_{1:t}, q_{1:t})$, the likelihood $\tilde{p}$, and the prior of class $c$, $\hat{p}'(c)$.
Using the likelihood model, the guesser selects a maximum a posterior solution for classifying the class.

\begin{equation}
\hat{p}(c|a_{1:t},q_{1:t}) \propto \hat{p}'(c) \prod_{\rt{j=1}}^{t} \tilde{p}(a_j|c,q_j,a_{1:j-1},q_{1:j-1})
\label{eq:eq_posterior}
\end{equation}

We use a term of likelihood as $\tilde{p}$, prior as $\hat{p}'$, and posterior as $\hat{p}$ from the perspective that the questioner classifies class $c$. 
During the conversation, $\hat{p}$ can be calculated in a recursive way.

\begin{algorithm}[t]
   \caption{AQM's Question-Generator}
   \label{alg:alg1}
\begin{algorithmic}
    \STATE $\hat{p}(c) \sim \hat{p}'(c)$-model
    \STATE $\tilde{p}(a_t|c,q_t,a_{1:t-1},q_{1:t-1}) \sim \tilde{p}(a|c,q)$-model
    \STATE $Q \leftarrow Q$-sampler
    \FOR {$t$ = 1:$T$}
       \STATE $q_t \leftarrow $ argmax$_{q_t \in Q}$ $\tilde{I}[C,A_t;q_t,a_{1:t-1},q_{1:t-1}]$ in Eq. \ref{eq:eq_QG}
       \STATE Get $a_t$ from the answerer
       \STATE Update $\hat{p}(c|a_{1:t},q_{1:t}) \propto \tilde{p}(a_t|c,q_t,a_{1:t-1},q_{1:t-1}) \cdot \hat{p}(c|a_{1:t-1},q_{1:t-1})$
    \ENDFOR
\end{algorithmic}
\end{algorithm}

\begin{figure}[t] 
\centering
\includegraphics[width=0.75\textwidth]{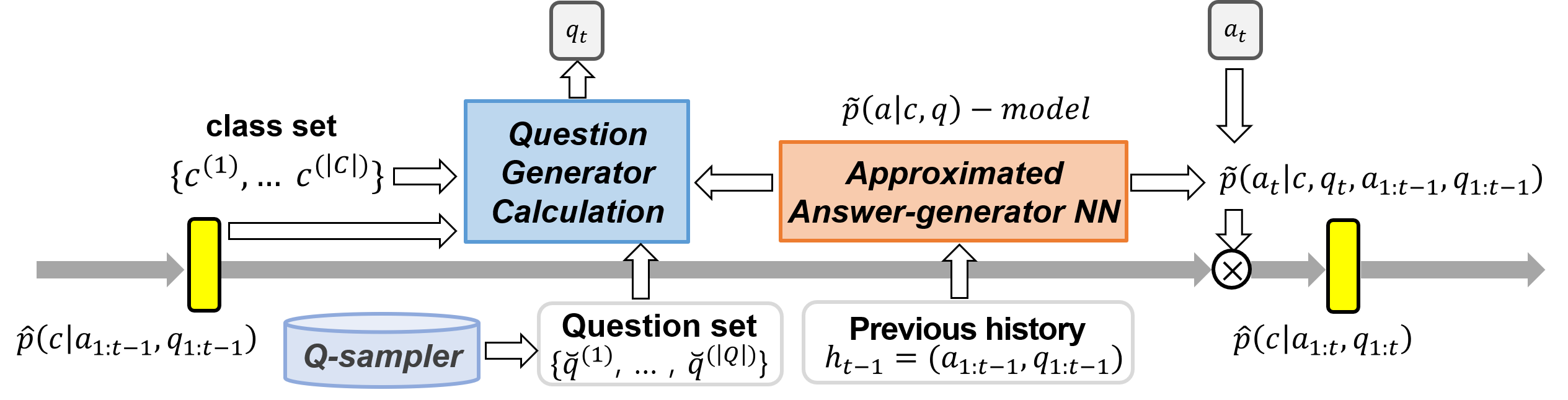}
\caption{Procedural illustration on the AQM's question-generator.}
\label{fig:icml18_fig3}
\end{figure}

The AQM's question-generator module selects $q^*_t$, which has a maximum value of the approximated information gain $\tilde{I}[C,A_t;q_t,a_{1:t-1},q_{1:t-1}]$, or simply $\tilde{I}$.
To calculate the information gain $\tilde{I}$, the question-generator module uses the likelihood $\tilde{p}$ and the posterior $\hat{p}$.

\begin{equation}
\begin{aligned}
q^*_t & = \argmaxA_{q_t \in Q} \tilde{I}[C,A_t;q_t,a_{1:t-1},q_{1:t-1}] \\
      & = \argmaxA_{q_t \in Q} \sum_{a_t} \sum_{c} \hat{p}(c|a_{1:t-1},q_{1:t-1})  \tilde{p}(a_t|c,q_t,a_{1:t-1},q_{1:t-1}) 
      \ln \frac{\tilde{p}(a_t|c,q_t,a_{1:t-1},q_{1:t-1})}{\tilde{p}'(a_t|q_t,a_{1:t-1},q_{1:t-1})}
\end{aligned}
\label{eq:eq_QG}
\end{equation}

\noindent
where $\tilde{p}'(a_t|q_t,a_{1:t-1},q_{1:t-1}) = \sum_{c} \hat{p}(c|a_{1:t-1},q_{1:t-1}) \cdot  \tilde{p}(a_t|c,q_t,a_{1:t-1},q_{1:t-1})$.
$Q$-sampler is required to \rt{select} the question from the candidate questions $Q$.
In our main experiments in Section 4.2, $Q$-sampler extracts candidate questions from the training data. In this case, AQM does not generate a new question for the test image.
However, if $Q$-sampler uses a RNN-based model, AQM can generate the question. We discuss this issue in Section 5.2.

\textbf{Learning }
In AQM, the answer-generator network in the questioner and the answerer does not share the representation. 
\rt{Thus, we need to train the AQM's questioner.}
In the existing deep learning framework, SL and RL are used to train two RNN-based models of the questioner.
In a typical deep SL method, questioner's RNNs are trained from the training data, which is the same or similar to the data the answerer is trained from.
In a typical deep RL method, the answerer and the questioner make a conversation in the self-play environment.
In this RL procedure, the questioner uses the answers generated from the answerer for end-to-end training, with reward from the result of the game.
On the other hand, the AQM's questioner trains the approximated answer distribution of the answerer, the likelihood $\tilde{p}$.
The likelihood $\tilde{p}$ can be obtained by learning training data as in deep SL methods, or using the answers of the answerer obtained in the training phase of the self-play conversation as in deep RL methods.
As the objective function of RL or AQM does not guarantee human-like question generation \cite{kottur2017}, RL uses SL-based pre-training, whereas AQM uses an appropriate $Q$-sampler.

Algorithm \ref{alg:alg1} and Figure \ref{fig:icml18_fig3} explain the question-generator module procedure. 
The question-generator requires the $\hat{p}'(c)$-model for the prior, the $\tilde{p}(a|c,q)$-model for the likelihood, and the $Q$-sampler for the set of candidate questions.
Additional explanations on AQM can be found in Appendix A.

\section{Experiments}
\subsection{MNIST Counting Dialog}

To clearly explain the mechanism of AQM, we introduce the MNIST Counting Dialog task, which is a toy goal-oriented visual dialog problem, illustrated in Figure \ref{fig:icml18_fig45} (Left). 
Each image in MNIST Counting Dialog contains 16 {small images of digit, each having four randomly assigned properties: color = \{red, blue, green, purple, brown\}, bgcolor = \{cyan, yellow, white, silver, salmon\}, number = \{0, 1, $\cdots$, 9\}, and style = \{flat,stroke\}. 
The goal of the MNIST Counting Dialog task is to inform the questioner to pick the correct image among 10K candidate images via questioning and answering. 
In other words, class $c$ is an index of the true target image (1 $\sim$ 10,000).

\begin{figure}[t] 
\centering
\includegraphics[width=0.45\textwidth]{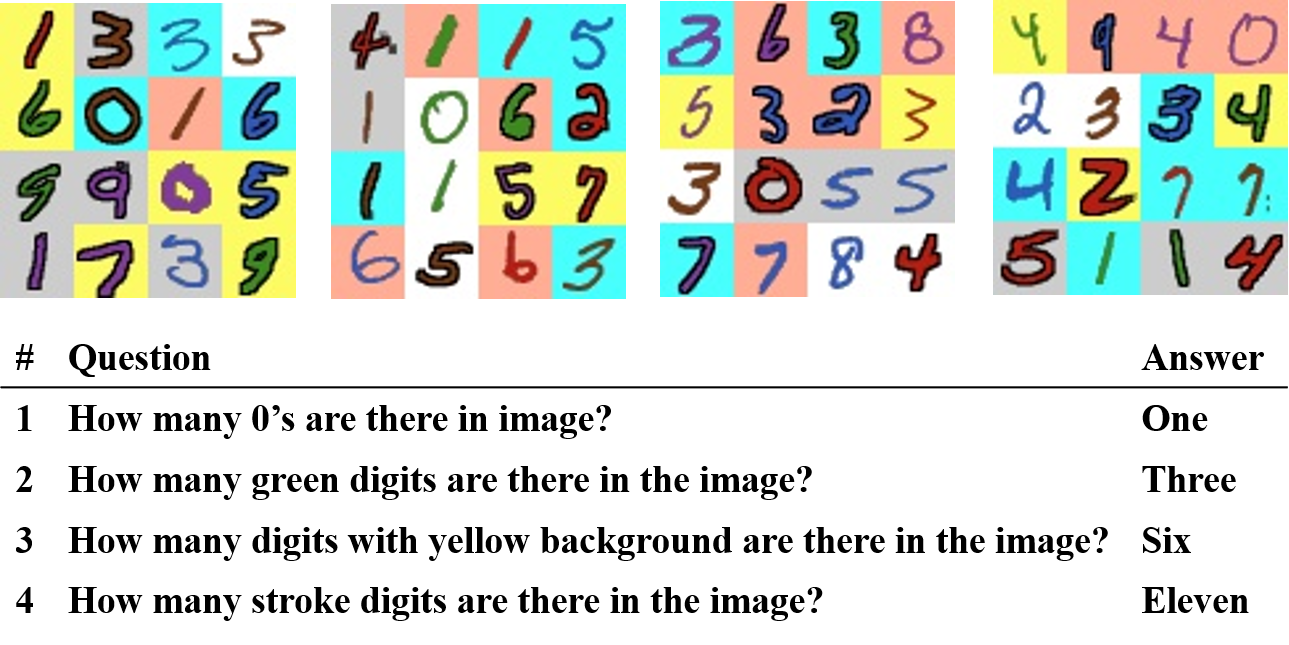}
\includegraphics[width=0.45\textwidth]{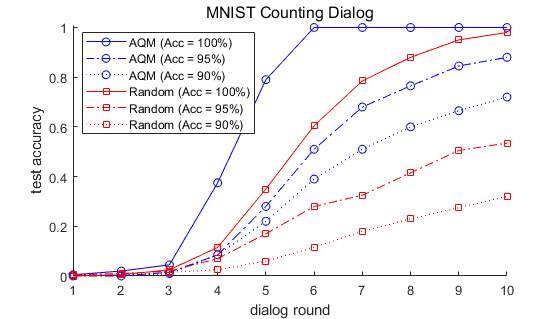}
\caption{(Left) Illustration of MNIST Counting Dialog, a simplified version of MNIST Dialog \cite{seo2017}. (Right) Test accuracy \ot{of goal-oriented dialog} from the MNIST Counting Dialog task. ``Acc'' is \ot{PropAcc, the average ratio of property recognition accuracy.}}
\label{fig:icml18_fig45}
\end{figure}

\bt{For the MNIST Counting Dialog task, we do not model the questioner and the answerer using neural networks. Instead, we define the answer model in the questioner is count-based on 30K training data. We set the average ratio of property recognition accuracy (PropAcc) $\lambda$} \ot{as 0.9, 0.95, and 1.0} \bt{in Figure 4. For four properties such as color, bgcolor, number and style, each property recognition accuracy $\lambda_{color}$, $\lambda_{bgcolor}$, $\lambda_{number}$, $\lambda_{style}$ is randomly sampled from an uniform distribution with a range of $\left[(2 \lambda-1), 1 \right]$. It contributes to add randomness or uncertainty to this task. For example, the percentage of correctly recognized color is 88\% if $\lambda_{color}$ is 0.88.}

\bt{According to the results in Figure \ref{fig:icml18_fig45}, if the PropAcc $\lambda$ decreases, the accuracy of goal-oriented dialogue which has a goal to select the correct image is decreased by a large amount.} Figure \ref{fig:icml18_fig45} (Right) shows that AQM nearly always chooses the true target image from 10K candidates in six turns if the \ot{PropAcc $\lambda$ is 1.0. However, AQM also chooses correctly with a probability of 51\% and 39\% (accuracy of goal-oriented dialog) in six turns, when the PropAcc $\lambda$ is 0.95 and 0.90, respectively. }
``Random'' denotes a questioner with a random question-generator module and the AQM's guesser module.
Detailed experimental settings can be found in Appendix B.

\subsection{GuessWhat?!}

\textbf{GuessWhat?! Task}
GuessWhat?! is a cooperative two-player guessing game proposed by De Vries et al. (Figure \ref{fig:icml18_fig12} (Right)) \cite{de2017}. 
GuessWhat?! has received attention in the field of deep learning and artificial intelligence as a testbed for research on the interplay of computer vision and dialog systems.
The goal of GuessWhat?! is to locate the correct object in a rich image scene by asking a sequence of questions.
One participant, ``Answerer'', is randomly assigned an object in the image. 
The other participant, ``Questioner,'' guesses the object assigned to the answerer. 
Both a questioner and an answerer sees the image, but the correct object is known only to the answerer.
To achieve the goal, the questioner asks a series of questions, for which the answerer responds as ``yes,'' ``no,'' or ``n/a.''
The questioner does not know a list of candidate objects while asking questions. 
When the questioner decides to guess the correct object, a list of candidate objects is then revealed. 
A win occurs when the questioner picks the correct object.
The GuessWhat?! dataset contains 66,537 MSCOCO images \cite{lin2014}, 155,280 games, and 831,889 question-answer pairs.

\noindent
\textbf{$\hat{p}'(c)$-model for the Prior }
The questioner does not know the list of candidate objects while asking questions. This makes the GuessWhat?! task difficult, although the number of candidates is around 8.
We use YOLO9000, a real-time object detection algorithm, to estimate the set of candidate objects \cite{redmon2016}.
The prior $\hat{p}'(c)$ is set to $1/N$, where $N$ is the number of extracted objects.

\noindent
\textbf{$\tilde{p}(a|q,c)$-model for the Likelihood }
We use the answerer model from previous GuessWhat?! research \cite{de2017}.
The inputs of the answer-generator module consist of a VGG16 feature of a given context image, a VGG16 feature of the cropped object in the context image, spatial and categorical information of the \ot{cropped} object, and the question $q_t$ at time step $t$.
A simple multi-layer perceptron model uses these features to classify the answer \{yes, no, n/a\}.
Our answer-generator module assumes the answer distribution is independent from the history $h_{t-1} = (a_{1:t-1},q_{1:t-1})$. In other words, we approximate the likelihood as $\tilde{p}(a_t|c,q_t,a_{1:t-1},q_{1:t-1}) \propto \tilde{p}''(a_t|c,q_t)$.

We use various strategy to train the questioner's approximated answer-generator network $\tilde{p}(a_t|c,q_t,a_{1:t-1},q_{1:t-1})$ to approximate the answerer's answer distribution $p(a_t|c,q_t,a_{1:t-1},q_{1:t-1})$.
In ``indA,'' $p$ and $\tilde{p}$ is trained separately for the same training data.
In ``depA,'' in which $\tilde{p}$ is trained for the answer inferred from the answerer $p$, where the question and the image is also sampled from the training data.
\rt{The performance improvement of indA and depA setting would be achieved partly because the answerer and the questioner share the dataset. For ablation study, we also introduce ``indAhalf'' and ``depAhalf'' setting.}
In \rt{indAhalf,} $p$ and $\tilde{p}$ is trained for the different dataset each, which the training data is exclusively divided into halves. In \rt{depAhalf,} $p$ is trained for the first half training data. 
After that, the questioner asks the questions in the second half training data to get the answer from $p$ and use the answer as the training label.

\noindent
\textbf{$Q$-sampler for the Candidate Question Set }
In the main experiments, we compare two $Q$-samplers which select the question from the training data. 
The first is ``randQ,'' which samples questions randomly from the training data.
The second is ``countQ,'' which causes every other question from the set $Q$ to be less dependent on the other. 
countQ checks the dependency of two questions with the following rule: the probability of that two \rt{sampled} questions’ answers are the same cannot exceed 95\%. 
In other words, $\sum_a \tilde{p}^\dagger (a_i=a|q_i,a_j=a,q_j) < 0.95$, where $\tilde{p}^\dagger (a_i|q_i,a_j,q_j)$ is derived from the count of a pair of answers for two questions in the training data.
$\tilde{p}$ made by indA is used for countQ.
We set the size of $Q$ to 200.

\begin{table}[t]
\begin{minipage}[t]{0.48\textwidth}
\centering
\captionof{figure}{Test accuracy from the GuessWhat?!. Previous works do not report the performance change with an increase in the number of turns.}
\includegraphics[width=1.00\textwidth]{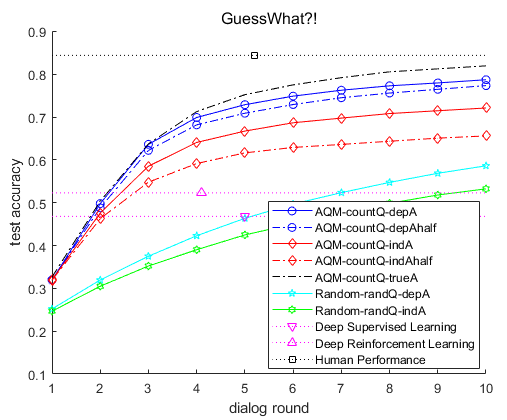}
\label{fig:icml18_fig6}
\end{minipage}\hfill
\begin{minipage}[t]{0.50\textwidth}
\centering
 \caption{Test \ot{accuracy} from the GuessWhat?!.}
    \begin{tabular}{ l  c }
    \\ \hline
    Model & Accuracy \\ \hline
    Baseline & 16.04  \\ \hline
    Deep SL (5-q) \cite{de2017} & 46.8  \\
    Deep RL (4.1-q in Avg) \cite{strub2017} & 52.3 \\ \hline
    \textbf{Random-randQ-indA (5-q)} & 42.48 ($\pm$ 0.84) \\ 
    \textbf{Random-randQ-depA (5-q)} & 46.36 ($\pm$ 0.91) \\ 
    \textbf{AQM-randQ-indA (5-q)} & 65.66 ($\pm$ 0.55) \\ 
    \textbf{AQM-countQ-indAhalf (5-q)} & 61.64 ($\pm$ 0.97) \\
    \textbf{AQM-countQ-indA (5-q)} & 66.73 ($\pm$ 0.76) \\
    \textbf{AQM-countQ-depAhalf (5-q)} & 70.90 ($\pm$ 1.14) \\
    \textbf{AQM-countQ-depA (5-q)} & 72.89 ($\pm$ 0.70) \\ \hline
    \textbf{AQM-countQ-depAhalf (10-q)} & 77.35 ($\pm$ 0.85) \\
    \textbf{AQM-countQ-depA (10-q)} & \textbf{78.72} ($\pm$ 0.54) \\ \hline
    Human  \cite{strub2017} & 84.4  \\ \hline
    \\
  \end{tabular}
  
  \label{table:table2}
\end{minipage}

\end{table}

\textbf{Experimental Results }
Figure \ref{fig:icml18_fig6} and Table \ref{table:table2} shows the experimental results. Figure \ref{fig:icml18_fig7} illustrates the generated dialog.
Our best algorithm, AQM-countQ-depA, achieved 63.63\% in three turns, outperforming deep SL and deep RL algorithms.
By allowing ten questions, the algorithms achieved 78.72\% and reached near-human performance.
If the answerer's answer distribution $p$ is directly used for the questioner's answer distribution $\tilde{p}$ (i.e., $\tilde{p}$ = $p$), AQM-countQ achieved 63.76\% in three turns and 81.96\% in ten turns (AQM-countQ-trueA in Figure \ref{fig:icml18_fig6}).
depA remarkably improved the score in self-play but did not increased the quality of the generated dialog significantly. 
On the other hand, countQ did not improve the score much but increased the quality of the generated dialog.
It is noticeable that the performance gap between indAhalf and indA is much larger than the gap between depAhalf and depA. This result shows that the conversation between the questioner and the answerer affects \ot{AQM's performance improvement} more than sharing the training data between the questioner and the answerer.
The \rt{compared} deep SL method used the question-generator with the hierarchical recurrent encoder-decoder \cite{serban2015}, achieving an accuracy of 46.8\% in five turns \cite{de2017}.
However, Random-randQ-depA achieved 46.36\% in five turns, which is a competitive result to the deep SL model.
``Random'' denotes random question generation from the randQ set.
The comparative deep RL method applied reinforcement learning on long short-term memory, achieving 52.3\% in about 4.1 turns \cite{strub2017}. 
The deep RL has a module to decide whether the dialog is stopped or not. 4.1 is the number of the averaged turns.

\section{Discussion}


\begin{figure}[t] 
\centering
\includegraphics[width=1.00\textwidth]{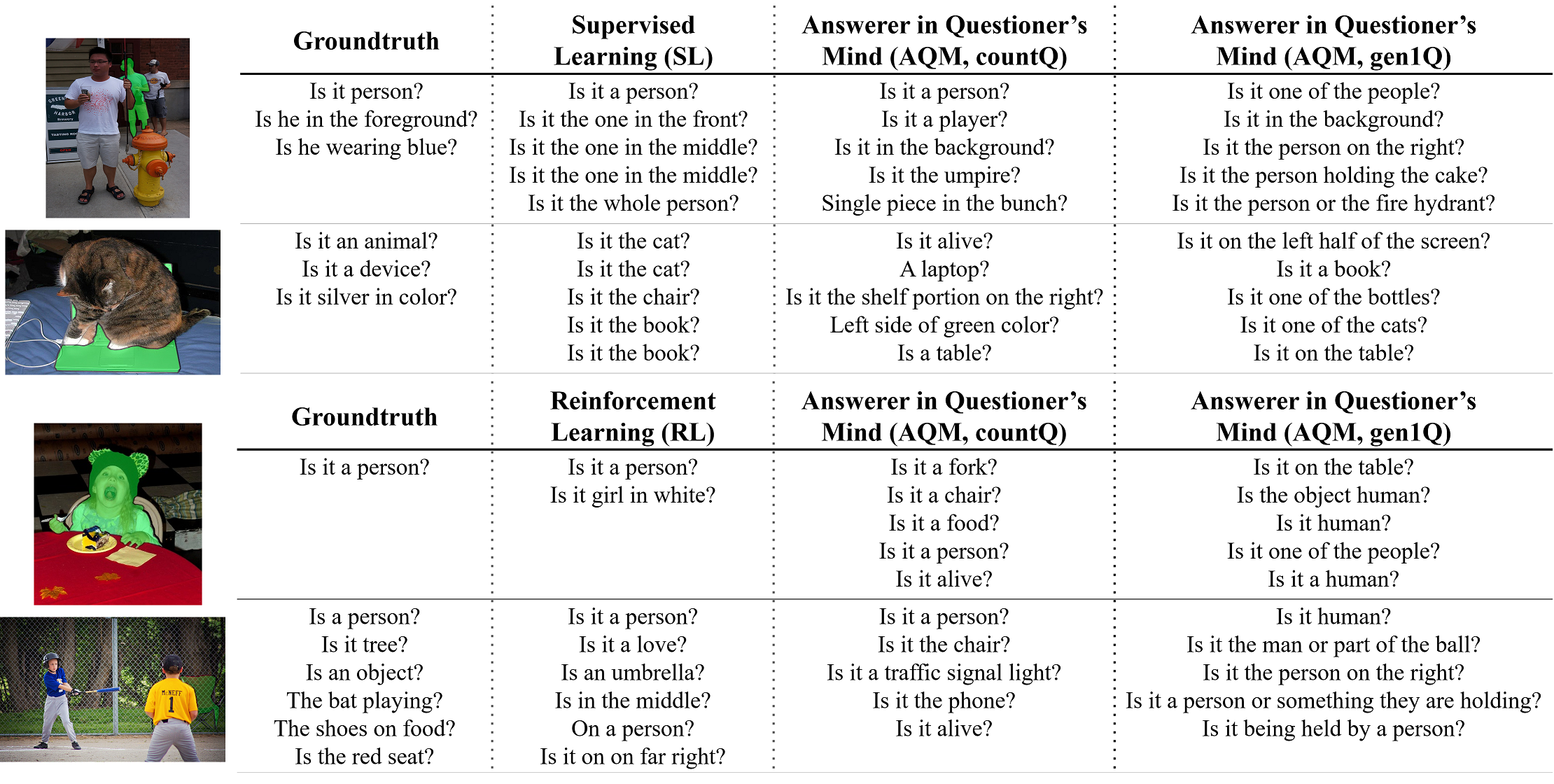}
\caption{Generated dialogs from our algorithm and the comparative algorithms. The tested games are sampled from the selected results of previous papers \cite{de2017,strub2017}.}
\label{fig:icml18_fig7}
\end{figure}

\subsection{Comparing AQM with SL and RL}

Our study replaced the task of training \rt{an} RNN which generates questions \rt{with} the task \rt{of} training a neural network which infers the probability of the \rt{answers.}
In the perspective of the hidden representation to track the dialog, the contexts of history which AQM's questioner requires are the posterior $\hat{p}$ and the history itself $h_t = (a_{1:t},q_{1:t})$ used as an input for the likelihood $\tilde{p}$.
In deep SL and RL methods, hidden neurons in RNN are expected to track the context of history.
If the question to be asked is independent from the previous questions, the only context AQM should track is the posterior $\hat{p}$. In this case, the posterior $\hat{p}$ in the yellow box of Figure \ref{fig:icml18_fig3} corresponds to the hidden vector of the RNN in the comparative dialog studies.

Moreover, we argue that, in Appendix D, AQM and RL have a similar objective function, just as information gain in decision tree is used to classify.
Many researchers have studied dialog systems for cooperative games using deep RL, to increase the score in self-play environments \cite{kottur2017,mordatch2017}.
In Appendix D, we leverage AQM as a tool for analyzing the deep RL approach on goal-oriented dialog tasks from the perspective of theory of mind.
According to our argument, training two agents to make plausible dialogs via rewards during self-play is not adaptable to a service scenario. To enable an agent to converse with a human, the opponent agent in self-play should model a human as much as possible. 
We prove that AQM and RL have a similar objective function, implying that RL-based training \ot{for} the questioner can also be seen as implicit approximation on the answer distribution of the answerer.

\subsection{Generating Questions}
For making a question more relevant to the new context of the dialog, a question needs to be generated.
Extracting a question from the training dataset (randQ and countQ) is just one of the settings the AQM's $Q$-sampler can have.
AQM can generate questions by using rule-based program \cite{rothe2017} or a seq2seq models previously used in goal-oriented dialog studies.
As a simple extension, we test the ``gen1Q'' setting, which uses a previous deep SL question-generator \cite{de2017} to generate a first-turn question for the test image.
We use a beam search to make a set of questions sorted by its likelihood, and select top-100 questions for the candidates.
In the experiments on GuessWhat?!, AQM-gen1Q-depA achieves a slight performance improvement over AQM-countQ-depA at 2-q (49.79\% $\rightarrow$ 51.07\%) outperforming the original deep SL method (46.8\% in 5-q).
However, at 5-q, AQM-gen1Q performs slightly worse than AQM-countQ-depA (72.89\% $\rightarrow$ 70.74\%).
If the Q-sampler generates questions through the seq2seq model using the history of dialog at every turn, the performance would be improved further.
Appendix E discusses the further direction of future works to make the question agent applicable for service.

Figure \ref{fig:icml18_fig7} shows the generated dialogs of AQM-gen1Q. gen1Q tends to make more redundant sentences than countQ because countQ carefully checks the similarity between questions. However, gen1Q tends to make the questions more related to the image. It is also noticeable that there are questions concatenating two sentences with ``or'' in gen1Q.
The score of the game is insufficient to evaluate the quality of the generated dialog. Appendix C discusses the objective function of the goal-oriented dialog, mainly based on the case of goal-oriented visual dialog studies.

\section{Conclusion}

We proposed \ot{``Answerer in Questioner's Mind'' (AQM),} a practical goal-oriented dialog framework using information-theoretic approach.
In AQM, the questioner approximates the answerer's answer distribution in dialog.
In our experiments, AQM outperformed deep SL and RL methods.
AQM can be implemented in various manners, not relying on a specific model representation nor a learning method. 
We also extended AQM to generate question by applying a previously proposed deep SL method. 
In this case, AQM can be understood as a way to boost the existing deep learning method.
Throughout the paper, we argued that considering the collaborator's mind in implementing an agent is useful and fundamental.

\section*{Acknowledgements}
The authors would like to thank Jin-Hwa Kim, Tong Gao, Cheolho Han, Wooyoung Kang, Jaehyun Jun,  Hwiyeol Jo, Byoung-Hee Kim, Kyoung Woon On, Sungjae Cho, Joonho Kim, Seungjae Jung, Hanock Kwak, Donghyun Kwak, Christina Baek, Minjoon Seo, Marco Baroni, and Jung-Woo Ha for helpful comments and editing.
\rt{This work was supported by the Institute for Information \& Communications Technology Promotion (R0126-16-1072-SW.StarLab, 2017-0-01772-VTT , 2018-0-00622-RMI) and Korea Evaluation Institute of Industrial Technology (10060086-RISF) grant funded by the Korea government (MSIP, DAPA).}

\small
\setlength{\bibsep}{5.0pt}
\bibliography{nips_2018}
\bibliographystyle{unsrt}
\normalsize

\newpage

\section*{Appendix A. Additional Explanations on AQM}

In our problem setting, three modules exist: an answer-generator, a question-generator and a guesser. 
The objective function of three modules is as follows.

\smallskip 
Answer-generator: $\mbox{argmax}_{a_t} \ p(a_t|c,q_t,a_{1:t-1},q_{1:t-1})$

Guesser: $\mbox{argmax}_{c} \ p(c|a_{1:t}, q_{1:t})$

Question-generator: $\mbox{argmax}_{q_t} \ I[C,A_t;q_t,a_{1:t-1}, q_{1:t-1}]$

\smallskip

If $a_t$ is a sentence, the probability of the answer-generator is extracted from the multiplication of the word probability of RNN.
For the objective function of question-generator, we use information gain $I$ of the class $C$ and the current answer $A_t$, where the previous history $(a_{1:t-1}, q_{1:t-1})$ and a current question $q_t$ are given.

\begin{equation}
\begin{aligned}
& I[C,A_t;q_t,a_{1:t-1},q_{1:t-1}] \\
= & H[C;a_{1:t-1},q_{1:t-1}] - H[C|A_t;q_t,a_{1:t-1},q_{1:t-1}] \\ 
= & - \sum_{c} p(c|a_{1:t-1},q_{1:t-1}) \ln p(c|a_{1:t-1},q_{1:t-1}) \\
& + \sum_{a_t} \sum_{c} p(c,a_t|q_t,a_{1:t-1},q_{1:t-1}) 
\ln  p(c|a_t,q_t,a_{1:t-1},q_{1:t-1}) \\
= & \sum_{a_t} \sum_{c} p(c,a_t|q_t,a_{1:t-1},q_{1:t-1}) 
\ln \frac{p(c|a_t,q_t,a_{1:t-1},q_{1:t-1})}{p(c|a_{1:t-1},q_{1:t-1})} \\
= & \sum_{a_t} \sum_{c} p(c|a_{1:t-1},q_{1:t-1}) p(a_t|c,q_t,a_{1:t-1},q_{1:t-1}) 
\ln \frac{p(a_t|c,q_t,a_{1:t-1},q_{1:t-1})}{p(a_t|q_t,a_{1:t-1},q_{1:t-1})}
\label{eq:eq_infogain_appendix}
\end{aligned}
\end{equation}

For the derivation of the last line, \gt{Bayes rule} is used. We also assume that $p(c|q_t,a_{1:t-1},q_{1:t-1}) = p(c|a_{1:t-1},q_{1:t-1})$, because asking a question itself does not affect the posterior of class $c$.

In calculating information gain $I$, AQM directly assigns each probability value to the equation of $I$, and \gt{sums out} $a_t$ and $c$.
However, all probability functions in Equation \ref{eq:eq_infogain_appendix} are the distribution from the answerer. 
Our approach is to approximate the answerer's answer distribution $p(a_t|c,q_t,a_{1:t-1},q_{1:t-1})$ to the likelihood $\tilde{p}(a_t|c,q_t,a_{1:t-1},q_{1:t-1})$.
We use the approximated information gain $\tilde{I}$ to select an adequate question.

\begin{equation}
\begin{aligned}
  & \tilde{I}[C,A_t;q_t,a_{1:t-1},q_{1:t-1}] \\
= & \sum_{a_t} \sum_{c} \hat{p}(c|a_{1:t-1},q_{1:t-1})  \tilde{p}(a_t|c,q_t,a_{1:t-1},q_{1:t-1}) 
      \ln \frac{\tilde{p}(a_t|c,q_t,a_{1:t-1},q_{1:t-1})}{\tilde{p}'(a_t|q_t,a_{1:t-1},q_{1:t-1})}
\end{aligned}
\label{eq:eq_QG_appendix}
\end{equation}

In calculating $\tilde{I}$, $\tilde{p}$ can be obtained by approximated answer distribution model.
However, there are other probability functions \gt{$\hat{p}$, $\hat{p}'$, and $\tilde{p}'$} beside the answer distribution.
\gt{These three} functions can be again calculated from $\tilde{p}$.
The posterior $\hat{p}$ can be calculated by the likeilhood $\tilde{p}$ and the prior $\hat{p}'$.

\begin{equation}
\hat{p}(c|a_{1:t},q_{1:t}) \propto \hat{p}'(c) \prod_{\rt{j=1}}^{t} \tilde{p}(a_j|c,q_j,a_{1:j-1},q_{1:j-1})
\end{equation}

During the conversation, the posterior can be calculated in a recursive way using $\hat{p}(c|a_{1:t},q_{1:t}) \propto \tilde{p}(a_t|c,q_t,a_{1:t-1},q_{1:t-1}) \cdot \hat{p}(c|a_{1:t-1},q_{1:t-1})$.
$\hat{p}'$ needs to be pre-defined. It can be a uniform distribution or other knowledge can be also applied to give adequate prior.
$\tilde{p}'$ can be again calculated with $\tilde{p}$ and $\hat{p}$.

\begin{equation}
\tilde{p}'(a_t|q_t,a_{1:t-1},q_{1:t-1}) = \sum_{c} \hat{p}(c|a_{1:t-1},q_{1:t-1})   \tilde{p}(a_t|c,q_t,a_{1:t-1},q_{1:t-1})
\end{equation}

When the answerer's answer distribution $p(a_t|c,q_t,a_{1:t-1},q_{1:t-1})$ is fixed, the questioner achieves an ideal performance when the likelihood $\tilde{p}$ is the same as $p(a_t|c,q_t,a_{1:t-1},q_{1:t-1})$.



\section*{Appendix B. MNIST Counting Dialog}

To clearly explain the mechanism of AQM, we introduce the MNIST Counting Dialog task, which is a toy goal-oriented visual dialog problem, illustrated in Figure \ref{fig:icml18_fig45} (Left). 
In the example of Figure \ref{fig:icml18_fig45} (Left), asking about the number of 1 digits or 6 digits classifies a target image perfectly if \gt{property recognition accuracy} on each digit in the image \gt{$\lambda_{number}$} is 100\%. Asking about the number of 0 digits does not help classify, because all images have one zero.
If \gt{$\lambda_{number}$} is less than 100\%, asking about the number of 1 digits is better than asking about the number of 6 digits, because the variance of the number of 1 digits is larger than that of the 6 digits.

An answering model in questioner is trained for 30K training data. 22 questions (for from red to stroke) and corresponding 22 answerer's answers are used for learning each instance of the training data.
Answering model in questioner is count-based. For true answer $a^{real}$ and answerer's answer $a^{feat}$ of the question $q_t$, the questioner's likelihood $\tilde{p}$ is as follow.

\begin{equation}
\begin{aligned}
\tilde{p}(a_t|c,q_t,a_{1:t-1},q_{1:t-1})
& \propto \tilde{p}''(a_t|c,q_t) \\
& = \frac{\#a^{feat}|a^{real}+\epsilon}{\#a^{real}+\epsilon'}
\end{aligned}
\end{equation}

This equation \rt{uses} an independence assumption.
$\#a^{real}$ is the number of $a^{real}$ for the training data where the answer for the class $c$ and the question $q_t$ is $a^{real}$.
$\#a^{feat}|a^{real}$ is the number of case where $a^{real}$ and $a^{feat}$ appears together. $\epsilon$ and $\epsilon'$ is the constant for normalization.
As the questioner uses the answer inferred from the answerer in the training phase, our setting in MNIST Counting Dialog corresponds to depA in GuessWhat?!.

\section*{Appendix C. Objective Function of Goal-Oriented Visual Dialog}

There have been several kinds of visual-language tasks including image captioning \cite{vinyals2015b} and VQA \cite{antol2015,mao2016}, and recent research goes further to propose multi-turn visual dialog tasks \cite{kim2017}. This section discusses the objective functions in research on goal-oriented dialog, mainly based on the case of goal-oriented visual dialog studies \cite{de2017,das2017b}.

\textbf{Visual Dialog \ } In Visual Dialog \cite{das2017a}, two agents also communicate with questioning and answering about the given MSCOCO image.
Unlike GuessWhat?!, an answer can be a sentence and there is no restriction for the dialog answerer. 
Das et al. used this dataset to make a goal-oriented dialog task, where the questioner guesses a target image from 9,627 candidates in the test dataset \cite{das2017b}. 
It is noticeable that, however, the questioner only uses the caption of the image, not the image itself when generating questions, because the goal of the task is to figure out the target image.
The dataset includes a true caption of each image achieving percentile ranks of around 90\%. 
In their self-play experiments, adding information via a dialog improved the percentile ranks to around 93\%, where the questioner and answerer were trained with deep SL and deep RL methods. This means that the models predict the correct image to be more exact than 93\% of the rest images in the test dataset.
\gt{Note that their models only improved around 3\% of the percentile rank at most, which implies that the caption information is more important than the dialog. There are also relevant issues on the performance which can be informed in the author's Github repository\footnote{https://github.com/batra-mlp-lab/visdial-rl}. According to their explanation, 0-th turn the percentile rank (only using the caption information) could be improved further by fine-tuning the hyperparameters, but in this case, the percentile rank did not increase much (around 0.3 of the percentile rank) during the conversation.}

\textbf{Objective 1: Score } 
In goal-oriented visual dialog research, the score is used as one of the main measurements of dialog efficiency.
However, a high score can be achieved via optimization over $p(c|a_{1:T},q_{1:T})$, which is the objective function of RL. 
In particular, the agent can achieve a high score \gt{although} the probability distribution of both the questioner and answerer is not bound to a human's distribution, even when human-like dialog is generated. 
For example, in GuessWhat?!, Han et al. showed that pre-defined questions about location can provide an accuracy of 94.34\% in five turns \cite{han2017}. Their methods divided an image evenly into three parts using two vertical or horizontal lines for three cases of answer \{yes, no, n/a\}. 
A natural language-based protocol can also be created using size, color, category, or other major properties of the object. 

\textbf{Objective 2: Service } 
One of the ultimate objectives of goal-oriented dialog research is to create an agent that can be used in a real service \cite{bordes2017}. However, successful reports have been limited. 
Chattopadhyay et al. reported the human-machine performance of the deep RL method in a study on Visual Dialog \cite{das2017b, chattopadhyay2017}.
In their method, questioner and answerer were both fine-tuned by RL. Thus, the answerer’s answering distribution differed from the training data. 
This RL method also used the objective function of deep SL as the regularizer, conserving generated dialog as human-like. 
Nevertheless, the RL algorithm deteriorated the score in the game with a human, compared to deep SL. 
The authors also assessed six measures of generated dialog quality: accuracy, consistency, image understanding, detail, question understanding, and fluency. However, the human subjects reported that the deep RL algorithm performed worse than the deep SL algorithm for all measures, except ``detail.'’

\textbf{Objective 3: Language Emergence } 
Plenty of research has recently been published on language emergence with RL in a multi-agent environment. 
Some studied artificial (i.e., non-natural) language \cite{evtimova2018,lazaridou2018}, whereas others attempted to improve the quality of generated natural dialog. 
One of the best progress found in the latter study is the improvement on the quality of a series of questions in a multi-turn VQA.
When deep RL is applied, the questioner generates fewer redundant questions than deep SL \cite{strub2017,das2017b}.
It can be understood that the questioner in these methods are optimized by both $p(q_t|a_{1:t-1},q_{1:t-1})$ and $p(c|a_{1:T},q_{1:T})$.
Deep RL methods use deep SL algorithms as a pre-training method \cite{strub2017}, or use the objective function of $p(c|a_{1:T},q_{1:T}) + \lambda \cdot p(q_t|a_{1:t-1},q_{1:t-1})$ \cite{das2017b}.
These studies focused on or achieved improvement of the questioner more than the answerer. It is because the answerer gets the objective function directly for each answer (i.e., VQA), whereas the questioner does not.

\section*{Appendix D. Analyzing RL via AQM}

\texttt{AQM's Property 1.} \textit{The performance of AQM's questioner is optimal, where the likelihood $\tilde{p}$ is equivalent to the answering distribution of the opponent $p$.}
For the guesser module, the performance of the guesser with the posterior $\hat{p}$ is optimal when $\tilde{p}$ is $p$.
The performance of question-generator also increases as the $\tilde{p}$ becomes more similar to the opponent, when the $\hat{p}$ is fixed.

\texttt{RL's Property 1.} \textit{AQM and RL approaches share the same objective function.} Two algorithms have the same objective function with the assumption that $q_t$ only considers the performance of the current turn. The assumption is used in the second line of the following equation.

\begin{equation}
\begin{aligned}
\argmaxA_{q_t} &  \ \max_{q_{t-}} \ln p(c|a_{1:T},q_{1:T})\\
\approx \argmaxA_{q_t} & \ \ln p(c|a_{1:t},q_{1:t}) \\
= \argmaxA_{q_t} & \ E \left[ \ln \frac{p(a_t|c,q_t,a_{1:t-1},q_{1:t-1})}{p(a_t|q_t,a_{1:t-1},q_{1:t-1})} \right] \\
=\argmaxA_{q_t}  & \ I[C,A_t;q_t,a_{1:t-1},q_{1:t-1}]
\end{aligned}
\end{equation}

$q_{t-}$ denotes \gt{$\{q_{1:t-1},q_{t+1:T}\}$}.
In the third line, $a_t \sim p(a_t|c,q_t,a_{1:t-1},q_{1:t-1})$, $c \sim p(c|a_{1:t-1},q_{1:t-1})$, and Bayes rule is used. 
The assumption in the second line can be alleviated via multi-step AQM, which uses $I[C,A_{t:t+k};q_{t:t+k},a_{1:t-1},q_{1:t-1}]$ as the objective function of the question-generator module. 
In the multi-step AQM, the optimal question cannot be selected in a greedy way, unlike \gt{original} AQM. 
The multi-step AQM needs to search in a tree structure; a Monte Carlo tree search \cite{coulom2006} can be used to find a reasonable solution.

The RL and AQM question-generator are closely related, as is the discriminative-generative pair of classifiers \cite{ng2002}.
AQM's question-generator and guesser module explicitly have a likelihood $\tilde{p}$, whereas the RL's modules do not have explicitly.
The properties of \gt{AQM including} optimal conditions and sentence dynamics can be extended to RL.
The complexity of RL's question-generator can be decomposed to tracking class posteriors $p(c|a_{1:t-1},q_{1:t-1})$ and history $(a_{1:t-1},q_{1:t-1})$ for multi-turn question answering.
For human-like learning, the context for language generation $p(q_t|a_{1:t-1},q_{1:t-1})$ is also required; $Q$-sampler corresponds to this context.

\texttt{RL's Property 2.} \textit{Optimizing both questioner and answerer with rewards makes the agent’s performance with human worse.} This is true, even when the process improves a score during self-play or uses tricks to maintain a human-like language generation. 
The property of self-play in a cooperative goal-oriented dialog task is different from the case of AlphaGo, which defeated a human Go champion \cite{silver2017}. 
For example, in GuessWhat?!, the reversed response of the answerer (e.g., ``no'’ for ``yes'’) may preserve the score in the self-play, but it makes the score in the human-machine game near 0\%.

According to AQM's Property 1, for the play of a machine questioner and a human answerer, the performance is optimal only if the approximated answerer's distribution $\tilde{p}$ of AQM's questioner is the same as the human's answering distribution. 
According to RL's Property 1, RL and AQM shares the optimality condition about the distribution of the opponent. 
Fine-tuning an answerer agent with a reward makes the distribution of the agent different from a human's.
Therefore, fine-tuning both agents decreases performance in service situations. 
The experiment with a human, studied by Chattopadhyay et al. explained in Appendix C, empirically demonstrates our claim.

\texttt{RL's Property 3.} 
\textit{An alternative objective function exists, which is directly applicable to each question.}
A reward can only be applied when one game is finished. 
If the goal of training is to make language emergence itself or to make an agent for service, two agents can communicate with more than just question-answering for back-propagation, such as sharing attention for the image \cite{kim2017}. 
Cross-entropy for the guesser of each round can be considered to replace the reward.
Information gain can also be used not only for AQM but also for alternative objectives of back-propagation.

\section*{Appendix E. Future Works}
\textbf{RL with Theory of Mind }
RL methods can be enhanced in a service scenario by considering the answering distribution of human. It is advantageous for the machine questioner to ask questions for which the human answer is predictable \cite{chandrasekaran2017}. In other words, a question having a high VQA accuracy is preferred. The model uncertainty of the questioner can also be measured and utilized with recent studies on Bayesian neural networks and uncertainty measures \cite{kendall2017}. Because the questioner has the initiative of dialog, the questioner does not need to necessarily learn the entire distribution of human conservation. The question, which the questioner uses frequently in self-play, can be asked more to a human. Then, the obtained question-answer pairs can be used for improving the answerer, like in active learning.

\textbf{\rt{Combining} Seq2seq with AQM}
RL optimizes $p(c|a_{1:t},q_{1:t})$ for the questioner in goal-oriented dialog. In the perspective of natural dialog generation, however, RL can be understood as that the questioner are optimized by both $p(q_t|a_{1:t-1},q_{1:t-1})$ and $p(c|a_{1:T},q_{1:T})$, as described in Appendix D.
On the other hand, the $Q$-sampler in AQM corresponds to regularizing with $p(q_t|a_{1:t-1},q_{1:t-1})$ in the deep RL approach.
If the $Q$-sampler in our experiment is replaced with seq2seq trained by a deep SL method, AQM can generate a question by optimizing both sentence probability from the seq2seq model and information gain.
AQM can use following terms in every turn to generate questions.

\begin{equation}
I[C,A_t;q_t,a_{1:t-1}, q_{1:t-1}] +\lambda \cdot \tilde{p}^{\dagger}(q_t|a_{1:t},q_{1:t})
\end{equation}

$\lambda$ is a balancing parameter. $\tilde{p}^{\dagger}$ is a probability distribution of language modeling of the seq2seq model. Comparing with the gen1Q setting in the discussion section, this algorithm could make the dialog more related to the history of dialog.
In this idea, AQM can be understood as a way to boost deep SL methods. 

\gt{\textbf{Online Learning } Fine-tuning on the model is required for a novel answerer, a non-stationary environment \cite{foerster2017}, or a multi-domain problem.  We think that fine-tuning on the answerer model would be more robust than on the question-generating RNN model, making AQM would have an advantage, because research on training an answerer model on VQA tasks has been more progressed than training an RNN for the questioner.
On the other hand, if the experiences of many users are available, model-agnostic meta learning (MAML) can be applied for few-shot learning \cite{finn2017}.}





\end{document}